\documentclass[10pt,twocolumn,letterpaper]{article}

\usepackage{iccv}              % To produce the CAMERA-READY version

\usepackage[accsupp]{axessibility}  % Improves PDF readability for those with disabilities.

\usepackage{microtype}
\usepackage[utf8]{inputenc} % allow utf-8 input
\usepackage[T1]{fontenc}    % use 8-bit T1 fonts
\usepackage{paralist} % Compacted versions of enumerate and itemize.
\usepackage{marvosym} % 常用符号, 特别是打勾、打叉
\usepackage{bm} % 实现字体的斜体加粗
\usepackage{amssymb}   % 黑板粗体 \mathbb, 德文尖角体 \mathfrak, \backsim
\usepackage{mathrsfs} % 花体 \mathscr
\usepackage{tabularray} % tblr or talltblr or longtblr environment
\UseTblrLibrary{booktabs}       % professional-quality tables，也就是三线表
\usepackage{tabularx} % 自动计算列宽的表格包
\usepackage[flushleft]{threeparttable} % 给表格项添加注释
\usepackage{colortbl} % 表格行列颜色 rowcolor
\usepackage{makecell} % 表格内容换行, Xhline
\usepackage{multirow} % 多行表格
\usepackage{nicefrac} % 斜着的分号
\usepackage[dvipsnames]{xcolor} % 字体颜色
\usepackage{soul}
\usepackage{bbm}

\usepackage{algorithm}
\usepackage{algorithmic}
\definecolor{iccvblue}{rgb}{0.21,0.49,0.74}
\usepackage[pagebackref,breaklinks,colorlinks,allcolors=iccvblue]{hyperref}

\def\paperID{11968} % *** Enter the Paper ID here
\def\confName{ICCV}
\def\confYear{2025}

\title{
DISTA-Net: Dynamic Closely-Spaced Infrared Small Target Unmixing
}

\author{Shengdong Han$^{{1}*}$ 
\and
Shangdong Yang$^{1}$\thanks{Equal contribution.}\\
\and
Yuxuan~Li$^2$\\  
\and
Xin~Zhang$^2$\\
\and
Xiang~Li$^{2,3}$~~~~~~~~Jian~Yang$^{2}$~~~~~~~~Ming-Ming Cheng$^{2,3}$~~~~~~~~Yimian~Dai$^{2,3}$\thanks{Corresponding author: yimian.dai@gmail.com.}  \\ 
$^1$ School of Computer Science, Nanjing University of Posts and Telecommunications \\
$^2$ VCIP, CS, Nankai University ~~~~~ $^3$NKIARI, Futian, Shenzhen
}

\begin{document}
\maketitle

% !TEX root = ../main.tex

\begin{abstract}
Resolving closely-spaced small targets in dense clusters presents a significant challenge in infrared imaging, as the overlapping signals hinder precise determination of their quantity, sub-pixel positions, and radiation intensities. 
While deep learning has advanced the field of infrared small target detection, its application to closely-spaced infrared small targets has not yet been explored. This gap exists primarily due to the complexity of separating superimposed characteristics and the lack of an open-source infrastructure.
In this work, we propose the Dynamic Iterative Shrinkage Thresholding Network (DISTA-Net), which reconceptualizes traditional sparse reconstruction within a dynamic framework.
DISTA-Net adaptively generates convolution weights and thresholding parameters to tailor the reconstruction process in real time.
To the best of our knowledge, DISTA-Net is the first deep learning model designed specifically for the unmixing of closely-spaced infrared small targets, achieving superior sub-pixel detection accuracy.
Moreover, we have established the first open-source ecosystem to foster further research in this field. This ecosystem comprises three key components: (1) CSIST-100K, a publicly available benchmark dataset; (2) CSO-mAP, a custom evaluation metric for sub-pixel detection; and (3) GrokCSO, an open-source toolkit featuring DISTA-Net and other state-of-the-art models, available at \url{https://github.com/GrokCV/GrokCSO}.
\end{abstract}

% \begin{IEEEkeywords}
% Infrared Small Target; Closely-Spaced Objects; 
% % Overlapping 
% Target Super-Resolution; Deep Unfolding; Attention Mechanism
% % ; HyperNetworks
% \end{IEEEkeywords}
% % \vspace{-1\baselineskip}
% !TEX root = ../main.tex
% \bibliography{../reference.bib}

\section{Introduction}
% 1. 研究背景的意义

Infrared imaging plays a pivotal role in various long-distance detection and surveillance tasks~\cite{TGRS2016TIRReview}, due to its exceptional sensitivity to thermal radiation and independence from illumination conditions.
However, the radiation intensity captured from remote targets is inherently weak due to their long-range distance from the imaging system~\cite{PR2023IRSTDSurvey}.
\textbf{This challenge is exacerbated when targets appear in spatially close dense clusters, as the closely-spaced objects (CSO)}~\cite{AMOS2023MuyGPyS}, \textbf{resulting in overlapping blob-like spots.}

\begin{figure*}[!t]
    \centering
    \includegraphics[width=.92\textwidth]{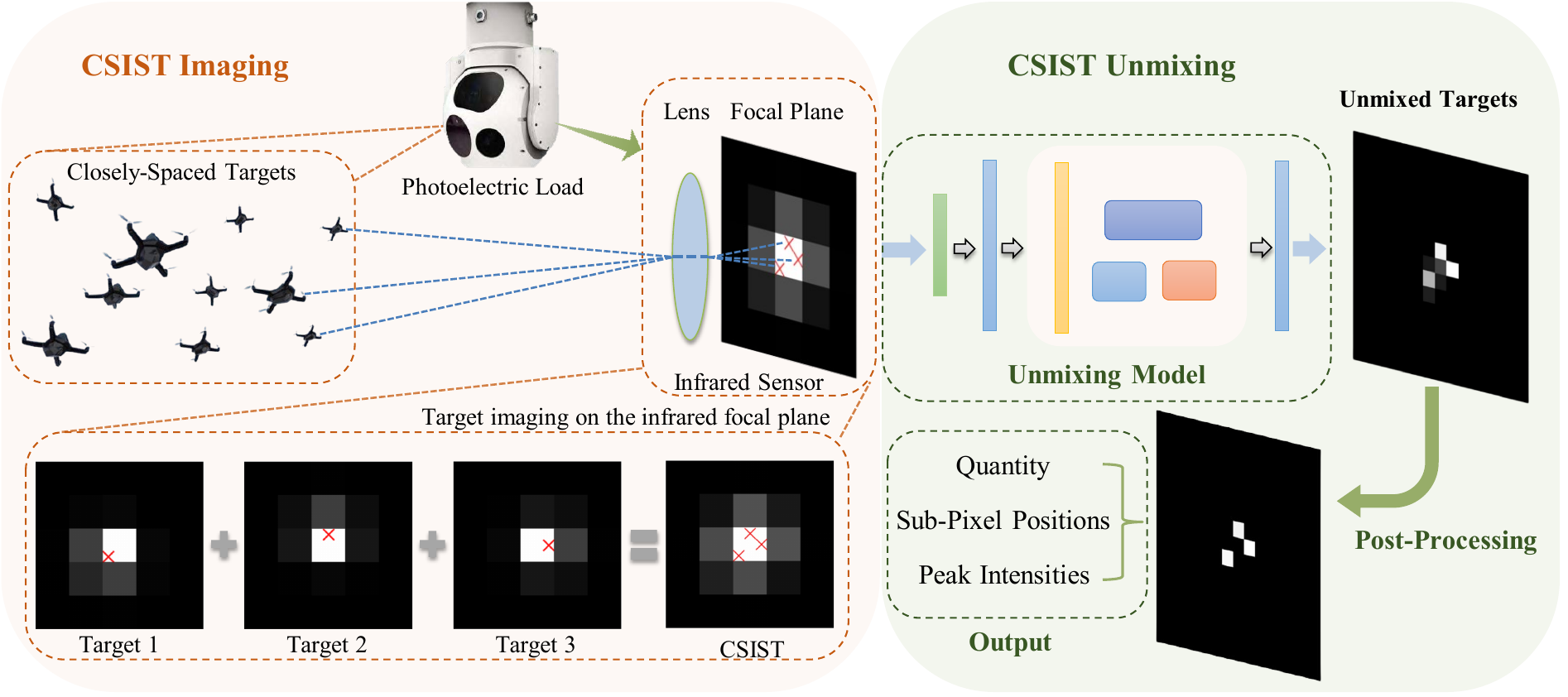}
    \caption{Conceptual illustration of imaging and unmixing processes for closely-spaced infrared small targets (CSIST). CSIST unmixing aims to disentangle and accurately estimate the count, positions, and intensities of overlapping targets.}
    \label{fig:CSIST-Imaging-Super-Resolution}
\end{figure*}

As shown in Fig.~\ref{fig:CSIST-Imaging-Super-Resolution}, such overlap makes it impossible to resolve the targets independently via human vision, thus obscuring the perception of target count, precise locations, and radiation intensities~\cite{GRSM2022SingleFrameSurvey}, presenting a substantial impediment for Infrared Search and Tracking (IRST) systems in their subsequent phases of detection, tracking, and identification.
Therefore, exploring effective techniques for unmixing and reconstruction of such \textbf{closely-spaced infrared small targets (CSIST)}, to accurately discern their exact locations and radiant intensities, holds great significance.

Despite the critical importance of CSIST unmixing in various applications, \textbf{research addressing this specific task remains exceedingly scarce}.
These approaches typically formulate the problem as a parameter estimation task and employ optimization algorithms to solve it~\cite{TSP2021ExtendedObjects}.
Target sparsity on the imaging plane was leveraged to devise a discretized sampling-based sparse reconstruction method~\cite{Zhang2013Sparse}, utilizing an over-complete dictionary and solving a second-order cone programming problem under the $\ell_1$ norm regularization.
However, the performance of these optimization-based models is highly dependent on meticulous hyperparameter tuning~\cite{NIPS2021Hyperparameter}, which poses significant challenges in real-world scenarios.
Variations in target quantity or location further complicate the selection of optimal hyperparameters, limiting the generalizability and practicality of these methods~\cite{TBDATA2017Hyperparameter}. Consequently, there is a pressing need for the development of unmixing algorithms that exhibit greater robustness to hyperparameter variations and can be effectively applied in diverse real-world settings.

While deep learning has revolutionized image super-resolution \cite{SPM2023Superresolution}, its application to CSIST unmixing remains unexplored, primarily due to fundamentally different task objectives and ecosystem limitations. Unlike generic super-resolution that enhances clarity through high-frequency detail restoration \cite{wu2024fully}, infrared small target unmixing requires precise estimation of overlapping targets' counts, locations, and radiation intensities - essentially mapping images to specific target attributes rather than conventional pixel-space super-resolution. This challenge is further compounded by the absence of standardized benchmark datasets, task-specific evaluation metrics, and open-source implementations, creating significant barriers to developing and comparing deep learning approaches in this specialized domain.

% 4. 研究目标/文章动机
To address the aforementioned challenges, in this paper,  we propose a novel deep unfolding network for CSIST unmixing, termed as the Dynamic Iterative Shrinkage Thresholding Network (DISTA-Net).
We reformulate the traditional sparse reconstruction approach into a dynamic deep learning framework, which adaptively generates convolution weights and thresholding parameters to tailor the reconstruction process in real time.
Distinct from prior methods, the parameters associated with the proximal mapping (nonlinear transforms and shrinkage thresholds) are dynamically adapted to the input data, rather than being hand-crafted or fixed after training.
\textit{To the best of our knowledge, this represents the first deep learning-based effort towards CSIST unmixing.}

Furthermore, we establish a comprehensive open-source ecosystem to facilitate research in this domain, including CSIST-100K, an open benchmark dataset comprising 100,000 pairs of CSIST images and exact annotations of location and radiation intensities; CSO-mAP, a custom evaluation metric inspired by the mean average precision (mAP) from object detection, calibrated to evaluate the quantity, spatial positioning, and radiation intensity of the unmixed infrared targets; and GrokCSO, an open-source PyTorch-based toolkit encapsulating our DISTA-Net alongside other state-of-the-art models, empowering researchers to effortlessly leverage the CSIST-100K dataset.

Our contributions can be categorized into \textbf{Four} aspects: 
1. We reformulate CSIST unmixing as an interpretable deep unfolding problem. To our knowledge, this is the first deep learning based attempt for this task. 
2. Our proposed DISTA-Net is a dynamic deep unfolding network, which adaptively generates convolution weights and thresholding parameters to tailor the reconstruction process conditioned on the input data. 
3. We establish the first open-source ecosystem for this task, including the CSIST-100K dataset, the CSO-mAP metric, and the GrokCSO toolkit. 
4. We provide a comprehensive analysis of our approach, validating the importance of dynamic deep unfolding and the effectiveness of the DISTA-Net in addressing the CSIST unmixing task.

% 1. We reformulate CSIST unmixing as an interpretable deep unfolding problem. To our knowledge, this is the first deep learning based attempt for this task.

% 2. Our proposed DISTA-Net is a dynamic deep unfolding network, which adaptively generates convolution weights and thresholding parameters to tailor the reconstruction process conditioned on the input data.

% 3. We establish the first open-source ecosystem for this task, including the CSIST-100K dataset, the CSO-mAP metric, and the GrokCSO toolkit.

% 4. We provide a comprehensive analysis of our approach, validating the importance of dynamic deep unfolding and the effectiveness of the DISTA-Net in addressing the CSIST unmixing task.

% \begin{compactenum}
%   \item We reformulate closely-spaced infrared small targets unmixing as an interpretable deep unfolding problem. To our knowledge, this is the first deep learning based attempt for this task.
%   \item Our proposed DISTA-Net is a dynamic deep unfolding network, which adaptively generates convolution weights and thresholding parameters to tailor the reconstruction process conditioned on the input data.
%   \item We establish the first open-source ecosystem for this task, including the CSIST-100K dataset, the CSO-mAP metric, and the GrokCSO toolkit.
%   \item We provide a comprehensive analysis of our approach, validating the importance of dynamic deep unfolding and the effectiveness of the DISTA-Net in addressing the CSIST unmixing task.
% \end{compactenum}

% !TEX root = ../main.tex
% \bibliography{../reference.bib}

\section{Related Work} \label{sec:related}

\subsection{Infrared Small Target Detection}
% Infrared small target detection has garnered significant research attention in recent years, driven by the development of open ecosystems, including a range of open-source datasets \cite{TGRS2021ALCNet,TGRS2023OSCAR,li2024sm3det}.

Driven by a range of open-source datasets \cite{TGRS2021ALCNet,TGRS2023OSCAR,li2024sm3det}, infrared small target detection has garnered significant research attention in recent years.
% Predominant approaches are categorized into sparse plus low-rank decomposition \cite{TIP2013IPI} and deep learning methods \cite{ICCV2019MDvsFA}
% Sparse and low-rank decomposition methods exploit the non-local self-correlation of background patches, assuming they belong to a singular subspace or low-rank subspace clusters \cite{JSTARS2017RIPT}. 
% These methods aim to separate targets from the background by decomposing the original imagery into a sparse target representation and a low-rank background representation \cite{JPROC2018RPCAReview}.
% However, real-world distractors that manifest as outliers amidst the background pose a significant challenge \cite{IPT2016WIPI}.
% Traditional methods rely solely on grayscale intensity or rudimentary handcrafted features such as central differences and entropy, struggling to semantically discern true targets from distractors \cite{DSP2021ZhouFei}. 
% Moreover, these methods are highly sensitive to hyperparameter configurations, requiring extensive manual tuning, which hinders their practical utility \cite{TPAMI2022TRPCA}
% Recently, publicly available datasets have transformed infrared small target detection into a supervised learning challenge \cite{TIP2023DNANet}.
Current research mainly focuses on developing multi-scale feature fusion models to counteract the scarcity of intrinsic target features \cite{TGRS2023MultiscaleMultilevel}.
Dai \emph{et al.} introduced an asymmetric contextual modulation module that bridges high-level semantics with low-level details using a bottom-up pathway via point-wise channel attention \cite{WACV2021ACM}.
Wang \emph{et al.} merged reinforcement learning with pyramid feature fusion and proposed a global context boundary attention module to mitigate localized bright noise \cite{TGRS23RLPGBNet}.
Cheng \emph{et al.} proposed a difference-aware attention module with a dual-temporal aggregation module for global feature learning and channel activation, and a difference-attention module for multi-scale detection via local correlations \cite{mei2023d2anet}.
Tong \emph{et al.} adopted an encoder–decoder structure, enhancing feature extraction with an atrous spatial pyramid pooling and a dual-attention module, and using multiscale labels to focus on target edges and internal features \cite{TGRS23MSAFFNet}.
Zhang \emph{et al.} proposed an infrared small target detection framework integrating visual-textual information via CLIP-prompted SAM adaptation with a denoising module, achieving enhanced generalization capability for the infrared domain \cite{zhang2025mirsam}.

Our work focuses on infrared small targets but differs in two key aspects. First, while infrared small target detection precedes our study, we prioritize CSIST unmixing, where detecting overlapped targets is crucial for sub-pixel localization and radiation intensity prediction. Second, our task goes beyond detection by enabling sub-pixel-level localization and radiative intensity estimation, providing finer target characterization than binary detection.

% Our work focuses on infrared small targets but differs from previous studies in two main ways. First, while infrared small target detection is a precursor to our study, our primary focus is on the CSIST unmixing. Detecting potentially overlapped targets is essential for subsequent tasks like sub-pixel localization and radiation intensity prediction. Second, our task extends beyond simple detection by precisely locating targets at the sub-pixel level and estimating their radiative intensities, offering a more detailed understanding of target characteristics compared to the previous binary detection approach.

% Our work, while focusing on infrared small targets like aforementioned studies, differs significantly in two key aspects:
% \begin{compactenum}
%     \item \textbf{Distinct Stages in the Pipeline:} Infrared small target detection serves as a precursor to our study's focus---the super-resolution of closely-spaced infrared small targets. Detecting potentially overlapped targets is a prerequisite for the subsequent tasks of sub-pixel target localization and radiation intensity prediction.
%     \item \textbf{Different Prediction Objectives:} Our task goes beyond mere detection by pinpointing exact sub-pixel locations and estimating radiative intensities. This provides a deeper understanding of target characteristics, surpassing the binary presence-or-absence approach of previous works.
% \end{compactenum}

\subsection{Deep Unfolding}

Deep unfolding, as delineated in \cite{SPM2021AlgorithmUnrolling}, originated in 2010 with the Learned Iterative Shrinkage-Thresholding Algorithm (LISTA) \cite{ICML2010LISTA}, which reinterprets ISTA \cite{Daubechies2004ISTA} as a fully connected feed-forward neural network. 
This approach generalizes effectively to new samples, achieving ISTA-like accuracy with fewer iterations.
Subsequent works, such as ADMM-Net \cite{NIPS2016ADMMNet}, have unfolded the steps of the Alternating Direction Method of Multipliers (ADMM) into a deep learning framework, thereby improving the accuracy and efficiency of MRI reconstruction through a compressive sensing model optimized via end-to-end discriminative training.
Likewise, ISTA-Net \cite{CVPR2018ISTANet} has adopted an end-to-end learning approach for proximal mapping, enhancing the performance of compressive sensing for natural image reconstruction.

Motivated by such advances, deep unfolding has found applicability in a variety of computer vision tasks.
Notably, Li \emph{et al.} transformed a generalized gradient-domain total variation algorithm into a deep interpretable network for blind image deblurring, delivering superior performance with learned parameters \cite{TCI2020EfficientDeblurring}.
For image super-resolution, Guo \emph{et al.} incorporated trainable convolutional layers into the Discrete Cosine Transform framework, effectively mitigating artifacts and enabling learning from limited data \cite{guo2019adaptive}.
Solomon \emph{et al.} unfolded robust principal component analysis into a deep network, improving the distinction between microbubble and tissue signals in ultrasound imaging \cite{TMI2020DeepUnfoldedRPCA}.

Unlike previous methods with static parameters \cite{CVPR2018ISTANet,ICME2021ISTANETPlusPlus},
our DISTA-Net dynamically adapts proximal mapping weights based on the input, enabling an adaptive reconstruction process that caters to varying scenarios.

\section{CSIST Benchmark, Metric, and Toolkit} \label{sec:ecosystem}

% In addressing the overlooked domain of Closely Spaced Infrared Small Target Super-Resolution (CSIST-SR), we present a tripartite framework to invigorate research and facilitate advancements. Our contributions include:
% \begin{compactenum}
% \item \textbf{CSIST-100K}: An open dataset for the CSIST-SR task.
% \item \textbf{CSO-mAP}: An evaluation metric designed to assess the performance of CSIST-SR algorithms.
% \item \textbf{GrokCSO}: 
% An open-source toolkit, designed to integrate seamlessly with CSIST-100K and support the implementation of state-of-the-art CSIST-SR algorithms.

% \end{compactenum}

\begin{figure}[htbp]
    \centering
    \includegraphics[width=8cm]{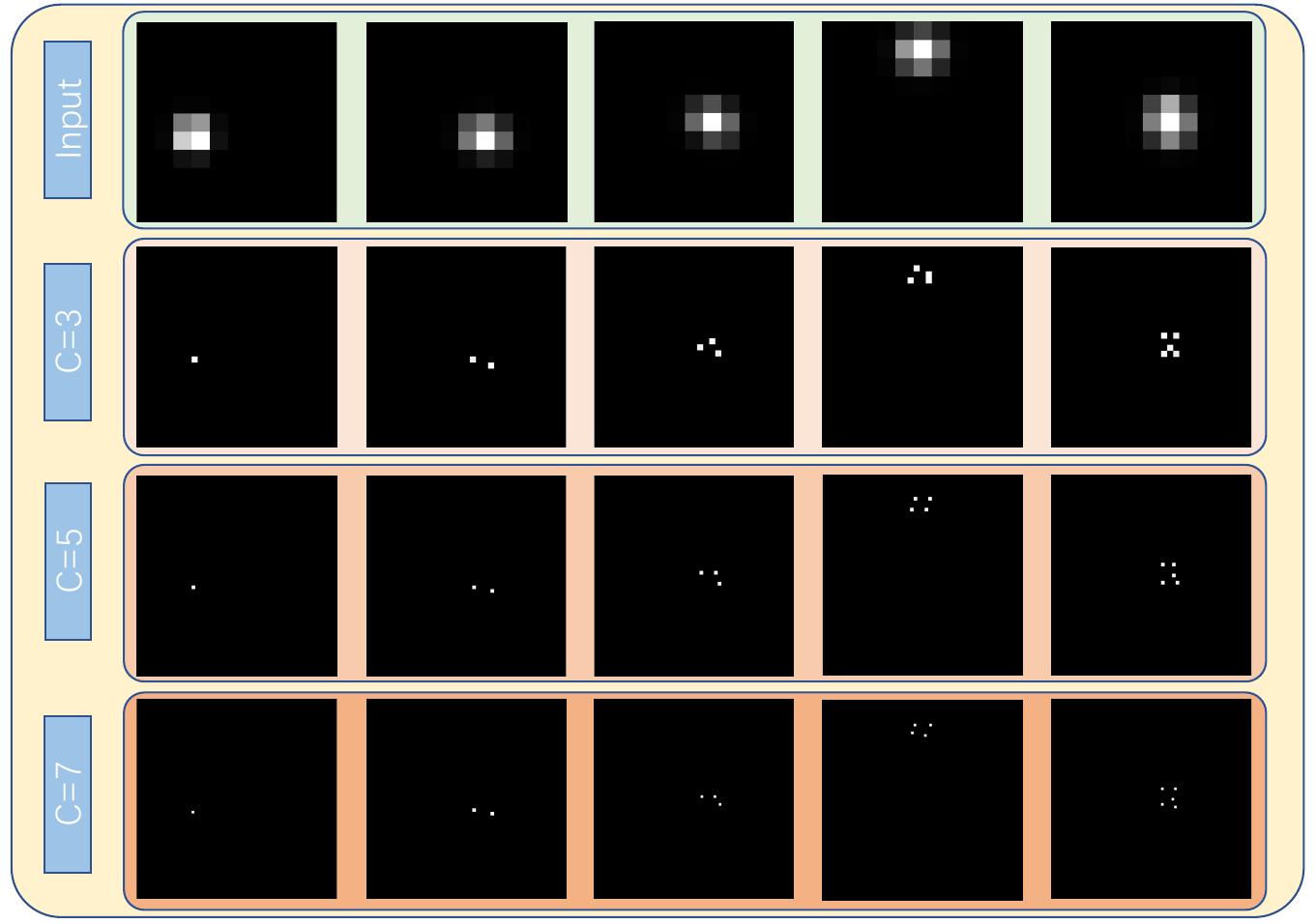}
    \caption{
    CSIST Visualization: The top row shows 1 to 5 overlapping targets, and the following rows display unmixing results for sub-pixel division factors of $3 \times$, $5 \times$, and $7 \times$.
    }
    \label{fig:gallery}
\end{figure}

\noindent \textbf{CSIST-100K Dataset.} \label{subsec:dataset}
% When a point light source is imaged through a lens, a diffraction pattern called the "Airy disk" forms at the focal point, featuring a bright central spot surrounded by concentric rings of alternating bright and dark fringes.
% This effect is approximated by the point spread function (PSF), whose standard deviation, dependent on the sensor's focal ratio ($f$-number) and detection band, quantifies the energy spread. In multi-target imaging,  each pixel's intensity is the cumulative response of overlaid point sources.
% % , as illustrated in Appendix ~\ref{appendix:Imaging}
% For remote objects, each acts as a point source, with the radius of the resultant Airy spot dictated by $1.22 \lambda/D$, where $\lambda$ is the wavelength and $D$ the lens diameter.
% This radius is equivalent to $1.9\sigma$ of a Gaussian PSF, defining the sensor's physical resolution limit per the Rayleigh criterion.
In our study, we set $\sigma_{\mathrm{PSF}}$ at 0.5 pixels. Simulations include 1–5 overlapping targets per image, each defined by 2D coordinates and radiation intensity (220–250 units), randomly placed within an $11 \times 11$ grid while maintaining $\geq$ 0.52 Rayleigh units separation. We generated the \textbf{CSIST-100K} dataset with 100,000 samples: 80,000 for training, and 20,000 split equally for validation and testing. As shown in Fig.~\ref{fig:gallery}, closely spaced targets diffuse, with energy concentrated in a $3 \times 3$ pixel area, causing significant overlap that complicates target counting and coordinate determination. (See \textbf{Supplementary} for optical modeling details.)

\setlength{\parskip}{0.5\baselineskip}
\noindent \textbf{CSO-mAP Metric.} \label{subsec:metric}
% Accurate localization of closely-spaced infrared small targets requires addressing the fundamental limitation of traditional bounding box metrics when inter-target distance falls below $1.9\sigma$ (due to Airy spot interference). CSO-mAP adapts the widely-used mAP framework from generic object detection but incorporates a specialized matching criterion that addresses the challenge of many-to-one matching problems. Our approach ensures each ground truth point is matched with at most one predicted point, prioritizing matches based on intensity values for accurate assessment. For more details, see Appendix ~\ref{appendix:metric}.
Traditional bounding box metrics fail when target separation falls below the Rayleigh criterion (due to Airy spot interference). Our CSO-mAP with strict sub-pixel position/intensity matching for CSIST evaluation, redefining TP/FP as follows:  
\begin{equation}
\mathbbm{1}_k\left(\hat{t}_j, t_i\right)=\left\{\begin{array}{lc}
1, & \text { if } d\left(\hat{t}_j, t_i\right) < \delta_k, \\
0, & \text { otherwise. }
\end{array}\right.
\end{equation}
Here, $\delta_k \in \left\{ 0.05, 0.1, 0.15, 0.2, 0.25\right\}$ is a series of distance thresholds, with $k = 1,2,3,4,5$, used to control the desired localization accuracy. Precision-recall (PR) curves are generated through intensity-prioritized matching, with Average Precision (AP) computed at each $\delta_k$. The final CSO-mAP metric is derived by averaging AP values across all thresholds, explicitly quantifying performance under varying spatial resolution demands (details in \textbf{ Supplementary}).

% The PR curve can be plotted by intensity prioritization to calculate the AP at different $\delta_k$, and the CSO-mAP can be obtained after averaging.

% Traditional bounding box metrics are inadequate for evaluating closely-spaced infrared targets, especially when targets' "Airy spots" overlap within $1.9\sigma$. We propose the closely spaced object mean Average Precision (CSO-mAP) metric for assessing localization accuracy and counting performance.
% To classify predictions as True Positive (TP) or False Positive (FP), we introduce a CSO-aware matching criterion. A predicted point $\hat{t}_j$ is considered TP if it matches an unmatched ground truth $t_i$:
% \begin{equation}
% \mathbbm{1}_k\left(\hat{t}_j, t_i\right)=\left\{\begin{array}{lc}
% 1, & \text { if } d\left(\hat{t}_j, t_i\right) < \delta_k \\
% 0, & \text { otherwise }
% \end{array}\right.
% \end{equation}
% where $\delta_k \in \left\{ 0.05, 0.1, 0.15, 0.2, 0.25\right\}$ are distance thresholds.
% Following COCO practices, we generate a Precision-Recall (PR) curve by varying confidence thresholds. The Average Precision (AP) is computed as the area under the PR curve. The final CSO-mAP is the average AP across all $\delta_k$, providing a standardized performance metric for CSIST-SR models.

\noindent \textbf{GrokCSO Toolkit.} \label{subsec:toolkit}
To address the lack of specialized tools in this domain, we introduce GrokCSO, an open-source toolkit for CSIST unmixing. Built on PyTorch, GrokCSO provides pre-trained models, reproducibility scripts, and specialized evaluation metrics tailored for CSO challenges. The detailed architecture and features of this toolkit are provided in \textbf{Supplementary}.
\section{Method} 
 In this section, we introduce the imaging model for closely-spaced infrared small targets, traditional sparse reconstruction approaches, and the proposed DISTA-Net architecture.

\subsection{Imaging and Unmixing Framework} \label{sec:framework}

\noindent \textbf{CSIST Imaging Model.}
Given the significant distance between targets and the infrared detector, targets can be approximated as point sources. The optical system's diffraction spreads the energy across adjacent pixels, described by a two-dimensional Gaussian point spread function (PSF) \cite{AA2021PSF}:
\begin{equation}
p(x, y)=\frac{1}{2 \pi \sigma_{\mathrm{PSF}}^2} \exp \left[-\frac{\left(x-x_t\right)^2+\left(y-y_t\right)^2}{2 \sigma_{\mathrm{PSF}}^2}\right] \text{, }
\label{eq:psf}
\end{equation}
where $\sigma_{\mathrm{PSF}}^2$ is the diffusion variance and $(x_t, y_t)$ the target coordinates.
On an infrared focal plane of $U \times V$ pixels, each pixel integrates the PSF within its boundaries:
\begin{equation}
g_{i, j}\left(x_t, y_t\right)=\int_{x_{i, j} -1 / 2 D}^{x_{i, j}+1 / 2 D} \int_{y_{i, j}-1 / 2 D}^{y_{i, j}+1 / 2 D} p(x, y) \mathrm{d} x \mathrm{~d} y,
\label{eq:response}
\end{equation}
where $(x_{i,j}, y_{i,j})$ is the pixel's center and $D$ the pixel width.
The focal plane measurement model is vectorized as:
\begin{equation}
\begin{aligned}
\mathbf{z}&=\mathbf{G}(\mathbf{x}, \mathbf{y}) \mathbf{s}+\mathbf{n},
\end{aligned}
\label{eq:measurement}
\end{equation}
where $\mathbf{G}(\mathbf{x}, \mathbf{y})$ is the steering matrix, $\mathbf{s}$ represents target intensities, and $\mathbf{n}$ denotes Gaussian white noise.

% For $N$ closely-spaced targets, each target is represented as $t_i=(x_i,y_i,g_i,\sigma_{\mathrm{PSF}})$, where $i \in {1,..,N}$, with $(x_i,y_i)$ denoting sub-pixel coordinates and $g_i$ the radiation intensities. The collections of center points and intensities are denoted  as $\mathcal{P}=\left\{p_i | i \in \{1,..,N\}\right\}$ and $\mathcal{G}=\left\{g_i | i \in \{1,..,N\}\right\}$.

\noindent \textbf{CSIST Unmixing via Sparse Reconstruction.}\label{subsec:proximal}
Given permissible quantization error, target positions in closely-spaced infrared small targets can be discretized into a finite set of sub-pixel locations $\Omega=\left\{(x_l, y_l)\right\}_{l=1, \cdots, L}$, where actual target positions form a sparse subset.

\begin{figure}[htbp]
\centering
\includegraphics[width=.42\textwidth]{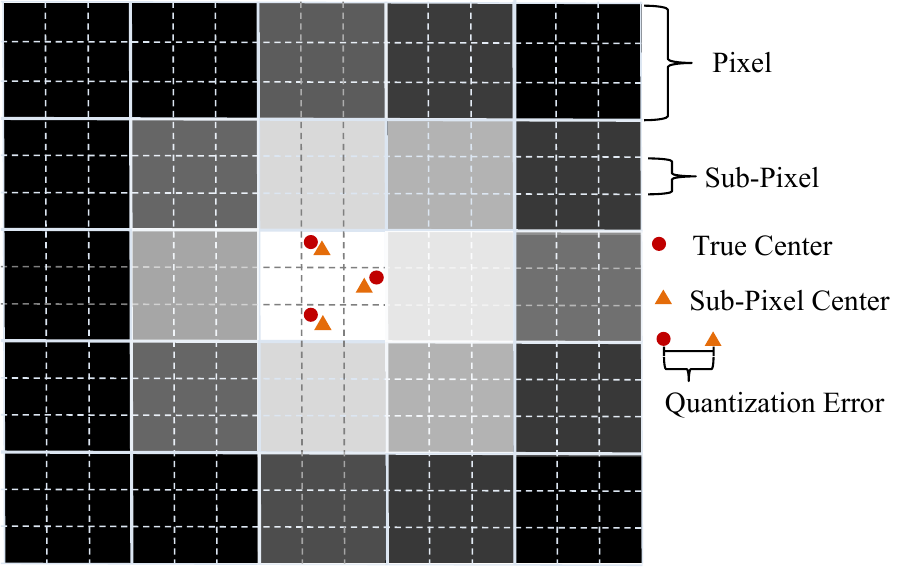}
\caption{Division of each pixel into an $n\times n$ grid of sub-pixels, representing potential target positions.}
\label{fig:subpixels}
\end{figure}

As shown in Fig. \ref{fig:subpixels}, each pixel is divided into an $n \times n$ grid, resulting in $L=UVn^2$ sub-pixels. With sufficient grid resolution, each sub-pixel contains at most one target, with maximum position deviation of $\sqrt{2}D/2n$.
The measurement model can be extended to $\Omega$:
\begin{equation}
\mathbf{z}=\mathbf{G}(\Omega) \tilde{\mathbf{s}}+\mathbf{w},
\label{eq:measurement2}
\end{equation}
where $\mathbf{G}(\Omega)$ contains steering vectors from $\Omega$, and $\tilde{\mathbf{s}} \in \mathrm{R}^{L}$ is sparse with $L \gg UV$.
The CSIST unmixing problem can then be formulated as a sparse reconstruction with $\ell_1$ regularization:
\begin{equation}
\min_{\tilde{\mathbf{s}}} \Vert \mathbf{z}-\mathbf{G}(\Omega) \tilde{\mathbf{s}}\Vert_2^2+\lambda\Vert\tilde{\mathbf{s}}\Vert_1,
\label{eq:reconstruction}
\end{equation}
where $\lambda$ is the regularization parameter.
The solution $\tilde{\mathbf{s}}$ directly yields target attributes: its non-zero entries indicate target count and intensities, while their corresponding positions in $\Omega$ provide sub-pixel coordinates.

% \noindent \textbf{Solution via Iterative Shrinkage-Thresholding Algorithm.}
\noindent \textbf{Optimization Solution.}
The ISTA \cite{Daubechies2004ISTA} solves the sparse reconstruction problem through two alternating steps:
\begin{align}
  \mathbf{r}^{(k)} &= \tilde{\mathbf{s}}^{(k-1)} - \rho \mathbf{G}^{\top}\left(\mathbf{G} \tilde{\mathbf{s}}^{(k-1)} - \mathbf{z}\right) ,\label{eq:update-r} \\
  \tilde{\mathbf{s}}^{(k)} &= \underset{\tilde{\mathbf{s}}}{\arg \min } \frac{1}{2}\left\|\tilde{\mathbf{s}} - \mathbf{r}^{(k)}\right\|_2^2 + \lambda\|\boldsymbol{\Psi} \tilde{\mathbf{s}}\|_1 \label{eq:update-x_1},
\end{align}
where $\boldsymbol{\Psi}$ denotes the transform matrix, $k$ the iteration index, and $\rho$ the step size.
The second step represents a proximal mapping:
\begin{equation}
  \operatorname{prox}_{\lambda \phi}(\mathbf{r})=\arg \min _{\tilde{\mathbf{s}}} \frac{1}{2}\|\tilde{\mathbf{s}}-\mathbf{r}\|_2^2+\lambda \phi(\tilde{\mathbf{s}}) .
  \label{eq:proximal}
\end{equation}
While ISTA with orthogonal transforms (e.g., wavelets) has efficient solutions, it faces challenges with complex transforms and requires numerous iterations. To address these limitations, ISTA-Net replaces $\mathbf{\Psi}$ with a trainable non-linear transform $\mathcal{F}(\cdot)$:
\begin{align}
  \tilde{\mathbf{s}}^{(k)} &= \underset{\tilde{\mathbf{s}}}{\arg \min } \frac{1}{2}\left\| \mathcal{F}\left( \tilde{\mathbf{s}} \right) - \mathcal{F}\left( \mathbf{r}^{(k)} \right)\right\|_2^2 + \theta\|\mathcal{F}\left( \tilde{\mathbf{s}} \right)\|_1 ,\label{eq:update-x_2}
\end{align}
where $\theta$ is a learnable parameter.
However, ISTA-Net's static network weights post-training limit its adaptability to input data, particularly in CSIST unmixing scenarios where input sensitivity is crucial.

% !TEX root = ../main.tex
% \bibliography{../reference.bib}
\begin{figure*}[htbp]
    \centering
    \includegraphics[width=.98\textwidth]{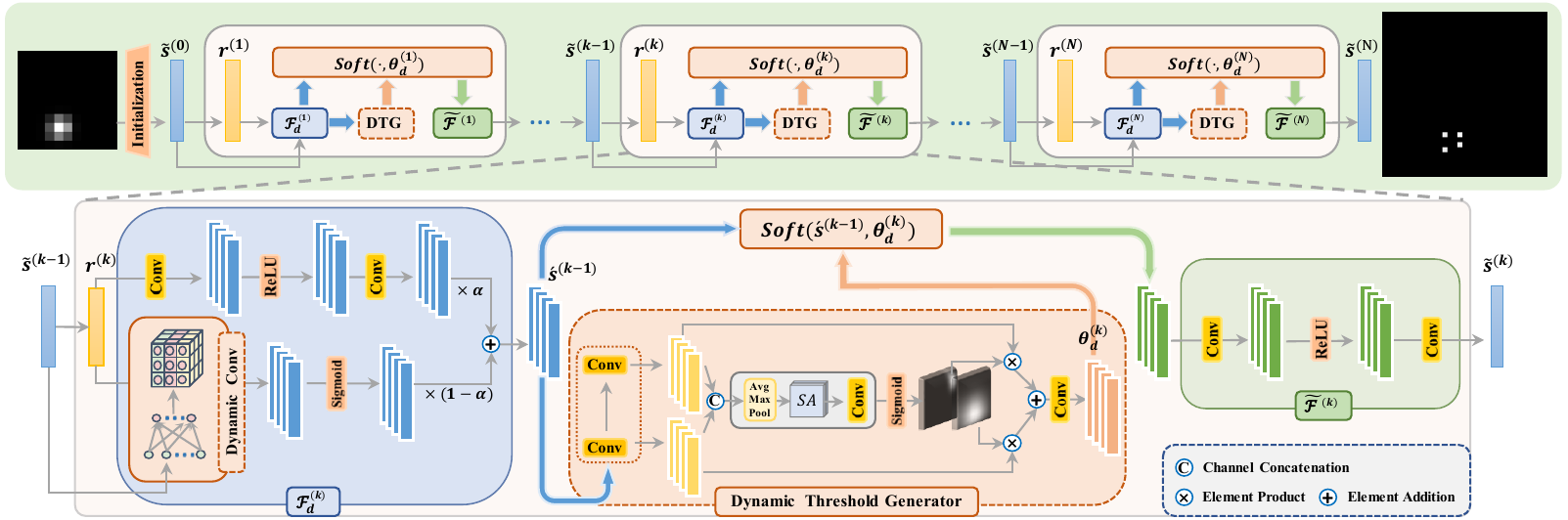}
    \caption{Architecture of the proposed DISTA-Net. The overall framework consists of multiple cascaded stages. Each stage contains three main components: a dual-branch dynamic transform module ($\mathcal{F}_d^{(k)}$) for feature extraction, a dynamic threshold module ($\Theta_d^{(k)}$) for feature refinement, and an inverse transform module ($\tilde{\mathcal{F}}^{(k)}$) for reconstruction.}
    \label{fig:Framework}
\end{figure*}

% \subsection{Dynamic ISTA Network}  \label{sec:unfolding}
\subsection{DISTA-Net: A Dynamic Framework}  \label{sec:unfolding}
\begin{algorithm}
\caption{DISTA-Net}
\label{alg:DISTA}
\begin{algorithmic}[1]  % 添加行号
\REQUIRE 
    CSIST image $\mathbf{z}$, steering matrix $\mathbf{G}(\mathbf{x}, \mathbf{y})$, initial matrix $Q_{\text{init}}$, number of stages $N$, step size $\{\rho^{(k)}\}_{k=1}^N$
\ENSURE 
    Reconstructed result $\tilde{\mathbf{s}}^{(N)}$
\\
\textbf{Learnable parameters:} 
\\
    $\{\rho^{(k)}\}_{k=1}^N$, 
    $\{\mathrm{DTG}^{(k)}\}_{k=1}^N$, 
    $\{\mathcal{F}_d^{(k)}\}_{k=1}^N$, 
    $\{\tilde{\mathcal{F}}^{(k)}\}_{k=1}^N$ \\
    ($\tilde{\mathcal{F}}^{(k)} \circ \mathcal{F}_d^{(k)} = \mathbf{I}$)
\\
\textbf{Initialization:}
\STATE $\tilde{\mathbf{s}}^{(0)} \gets Q_{\text{init}} \mathbf{z}$
\\
\textbf{Iterative reconstruction:}
\FOR{$k = 1$ \TO $N$}
    \STATE $\mathbf{r}^{(k)} \gets \tilde{\mathbf{s}}^{(k-1)} - \rho^{(k)} \mathbf{G}^\top (\mathbf{G} \tilde{\mathbf{s}}^{(k-1)} - \mathbf{z})$
    \STATE $\acute{\mathbf{s}}^{(k)} \gets \mathcal{F}_d(\tilde{\mathbf{s}}^{(k-1)}, \mathbf{r}^{(k)})$  
    \STATE $\theta_d^{(k)} \gets \mathrm{DTG}^{(k)}(\acute{\mathbf{s}}^{(k)})$
    \STATE $\tilde{\mathbf{s}}^{(k)} \gets \tilde{\mathcal{F}}(\operatorname{Soft}(\mathcal{F}_d(\mathbf{r}^{(k)}), \theta_d^{(k)}))$
\ENDFOR
\end{algorithmic}
\end{algorithm}
The details of our DISTA-Net are illustrated in Fig. ~\ref{fig:Framework}.
Building upon ISTA-Net's framework, we introduce two key improvements. First, we design a data-adaptive nonlinear transformation function $\mathcal{F}_d(\cdot)$ that maps images to richer dimensional representations while emphasizing significant image regions. The transformation follows:
\begin{equation}
  \mathcal{F}_d(\tilde{\mathbf{s}}^{(k)})=\operatorname{Soft}(\mathcal{F}_d(\mathbf{r}^{(k)}),\theta_d),
  \label{eq:F()}
\end{equation}
where $\operatorname{Soft}(\cdot,\theta_d)$ denotes soft-thresholding with learnable parameter $\theta_d$, and $k$ is the stage index. The left inverse $\tilde{\mathcal{F}}$ satisfies $\tilde{\mathcal{F}}(\cdot) \circ \mathcal{F}_d(\cdot) = \mathbf{I}$, without requiring structural symmetry to $\mathcal{F}_d(\cdot)$, yielding:
\begin{equation}
  \tilde{\mathbf{s}}^{(k)} = \tilde{\mathcal{F}}(\operatorname{Soft}(\mathcal{F}_d(\mathbf{r}^{(k)}), \theta_d)).
  \label{eq:s()}
\end{equation}
Second, we introduce a dynamic threshold module that adapts to input image variations, addressing the sensitivity of sparse vector perturbations in image generation. This flexible thresholding mechanism improves upon fixed parameters that can be either too strict or too lenient.
As shown in Fig.~\ref{fig:Framework}, DISTA-Net comprises $N$ stages. Each stage $k$ ($k>1$) contains three components: $\mathcal{F}_d^{(k)}$, $\operatorname{Soft}(\cdot, \theta_d^{(k)})$, and $\tilde{\mathcal{F}}^{(k)}$. The input $\mathbf{r}^{(k)}$ is derived from $\tilde{\mathbf{s}}^{(k-1)}$ via Eq. (\ref{eq:update-r}), processed through these components sequentially to generate $\tilde{\mathbf{s}}^{(k)}$ for the next iteration. Overall, the proposed DISTA-Net can be
 referenced in Algorithm ~\ref{alg:DISTA}.

\noindent \textbf{Dynamic Transform.}
While trainable non-linear transformations overcome limitations of handcrafted methods, their fixed post-training parameters result in static transformation patterns. We address this by introducing a dual-branch \textit{Dynamic Transform} module $\mathcal{F}_d^{(k)}$ at the $k$-th stage.

To handle the sensitivity of sparse image $\tilde{\mathbf{s}}^{(k-1)}$ perturbations, we design an auxiliary branch that guides $\mathbf{r}^{(k)}$ through a dynamic convolutional kernel. This approach enhances feature representation adaptively.

The module first processes $\tilde{\mathbf{s}}^{(k-1)}$ through a fully connected network to generate a weight vector:
\begin{equation}
  W = f(\tilde{\mathbf{s}}^{(k-1)}).
  \label{eq:W()}
\end{equation}
The \textit{Dynamic Conv} module applies this weight vector as an adaptive convolutional kernel to $\mathbf{r}^{(k)}$:
\begin{equation}
  w_r = C(W, \mathbf{r}^{(k)}).
  \label{eq:w()}
\end{equation}
The final output combines a Conv-ReLU-Conv branch with the sigmoid-activated auxiliary branch:
\begin{equation}
  \mathcal{F}_d^{(k)} = \alpha\cdot A(\text{ReLU}(B(\mathbf{r}^{(k)}))) + (1-\alpha)\cdot \text{sigmoid}(w_r),
  \label{eq:F()2}
\end{equation}
\noindent where $A(\cdot)$ and $B(\cdot)$ are convolution operations and $\alpha \in \left [ 0,1 \right ] $ governs the contribution between the two branches.

\setlength{\parskip}{0.5\baselineskip}
\noindent \textbf{Dynamic Soft-Thresholding.}
Unlike ISTA-Net's fixed threshold $\theta$, we propose a \textit{Dynamic Thresholding Generator(DTG)} module that adapts $\theta_d$ based on input image information. This approach better handles densely overlapped targets and spatial context variations.

As shown in Fig.~\ref{fig:Framework}, the module employs dual convolutional layers to capture multi-scale features as in~\cite{Li_2024_IJCV,sknet}. The $\mathcal{F}_d^{(k)}$ output passes through two $3\times 3$ convolutions, generating feature maps $\tilde{U}_1$ and $\tilde{U}_2$. These are concatenated to form $\tilde{U} = [\tilde{U}_1, \tilde{U}_2]$. Spatial relationships are captured through parallel pooling operations:
\begin{equation}
    SA_{\text{avg}} = P_{\text{avg}}(\tilde{U}), \quad 
    SA_{\text{max}} = P_{\text{max}}(\tilde{U}).
  \label{eq:SA}
\end{equation}
The pooled features are processed through the convolution:
\begin{equation}
    (\hat{SA}) = \text{Conv}^{2 \to N}([\text{SA}_{\text{avg}}; \text{SA}_{\text{max}}]),
    \label{eq:SA2}
\end{equation}
followed by a sigmoid activation to generate spatial selective masks:
\begin{equation}
    (\tilde{SA})_i = \text{sigmoid}((\hat{SA})_i).
    \label{eq:SA3}
\end{equation}
The dynamic threshold $\theta_d$ is then computed by combining these masks with multi-scale feature maps through a final convolution $C(\cdot)$:
\begin{equation}
    \theta_d = C(\sum_{i=1}^{N}{(\tilde{SA})_i \cdot \tilde{U}_i} ).
    \label{eq:theta}
\end{equation}

\paragraph{Initialization and Learning Objectives.}
DISTA-Net employs linear initialization similar to iterative sparse coding algorithms, where the initial estimate $\tilde{s}^{(0)}$ is obtained through an optimal linear projection $Q_{\text{init}}$ learned from training pairs $\{(z_i, s_i)\}$ (details in \textbf{ Supplementary}). 

The training objective combines reconstruction fidelity with structural preservation:
\begin{equation}
    \mathcal{L} = \mathcal{L}_{\text{discrepancy}} + \gamma \mathcal{L}_{\text{constraint}},
    \label{eq:L}
\end{equation}
where:
\begin{equation}
    \mathcal{L}_{\text{discrepancy}} = \frac{1}{M N_s} \sum_{i=1}^M \| \tilde{s}_i^{(N)} - s_i \|_2^2,
    \label{eq:Loss1}
\end{equation}
\begin{equation}
    \mathcal{L}_{\text{constraint}} = \frac{1}{M N_s} \sum_{i=1}^M \sum_{k=1}^N \| \tilde{\mathcal{F}}^{(k)} (\mathcal{F}_d^{(k)} (s_i)) - s_i \|_2^2.
    \label{eq:Loss2}
\end{equation}
Here, $\mathcal{L}_{\text{discrepancy}}$ computes the MSE between reconstructed $\tilde{s}_i^{(N)}$ and ground truth $s_i$, while $\mathcal{L}_{\text{constraint}}$ enforces multi-stage identity constraints through $\tilde{\mathcal{F}}^{(k)} \circ \mathcal{F}_d^{(k)} \approx \mathbf{I}$, with $\gamma=0.01$ balancing these objectives.

% !TEX root = ../main.tex
% \bibliography{../reference.bib}

\section{Experiments}
\label{sec:experiment}
% In this section, we validate the proposed DISTA-Net algorithm on the CSIST-100K dataset and compare its performance with state-of-the-art methods. We also present ablation studies and hyperparameter analysis to assess the effectiveness of each module and the model's sensitivity to parameters.

\begin{table*}[htbp]
  \renewcommand\arraystretch{1.2}
  \footnotesize
  \centering
    \vspace{-1.5\baselineskip}
  \vspace{-2pt}
  \setlength{\tabcolsep}{6.pt}
  \begin{tabular}{l|c|c|c|ccccc|c|c}
  \multirow{2}{*}{Method}   & \multirow{2}{*}{\#P $\downarrow$} & \multirow{2}{*}{FLOPs $\downarrow$} & \multicolumn{6}{c|}{CSO-mAP}  & \multirow{2}{*}{PSNR $\uparrow$} & \multirow{2}{*}{SSIM $\uparrow$}  \\
   & &  & mAP & AP-05  & AP-10    & AP-15    & AP-20  & AP-25 & & \\
  
  \Xhline{0.8pt}
  \multicolumn{9}{l}{\textit{Traditional Optimization}}  \\ \hline
  ISTA~\cite{Daubechies2004ISTA} & - & - &  7.46 & 0.01 & 0.31 & 2.39 & 9.46 & 25.14 & - & - \\
  \hline
  \multicolumn{9}{l}{\textit{Image Super-Resolution}}  \\ \hline
   ACTNet~\cite{zhang2023actnet} & 46.212M & 62.80G & 45.61 & 0.38 & 7.46 & 41.13 & 83.12 & 95.95 & 36.8879 & 0.9854 \\
  \hline
  CTNet~\cite{wang2021contextual} & 0.400M & 2.756G & 45.11 & 0.38 & 7.53 & 40.39 & 82.11 & 95.14 & 35.6154 & 0.9833 \\
  \hline
  DCTLSA~\cite{10215496} & 0.865M &13.69G& 44.51 & 0.39 & 7.35 & 39.35 & 81.15 & 94.34 & 35.2259 & 0.9832 \\
  \hline
  EDSR~\cite{lim2017enhanced} & 1.552M & 12.04G & 45.32 & 0.33 & 7.07 & 40.58 & 83.24 & 95.41 & 36.4349 & 0.9816 \\
  \hline
  EGASR~\cite{qiu2023cross} & 2.897M & 17.73G & 45.51 & 0.42 & 8.03 & 41.32 & \underline{85.71} & 95.08 & 35.4549 & 0.9745 \\
  \hline
  FeNet~\cite{wang2022fenet} & 0.348M & 2.578G & 45.77 & 0.42 & \underline{8.19} & 42.13 & 83.30 & 94.80 & 36.4507 & 0.9864 \\
  \hline
  RCAN~\cite{zhang2018image} & 1.079M & 8.243G & 45.87 & 0.42 & 7.96 & 41.81 & 83.61 & 95.57 & 36.5860 & 0.9831 \\
  \hline
  RDN~\cite{zhang2018residual} & 22.306M & 173.0G & 45.81 & 0.35 & 7.11 & 41.07 & 84.07 & 96.43 & 36.1883 & 0.9840 \\
  \hline
  SAN~\cite{dai2019second} & 4.442M & 34.05G & 45.95 & 0.36 & 7.35 & 41.17 & 84.32 & \underline{96.57} & 37.1805 & 0.9848 \\
  \hline
  SRCNN~\cite{dong2015image} & 0.019M & 1.345G & 29.06 & 0.23& 4.10 & 21.65 &  49.95 & 69.39 & 29.4140 & 0.9566 \\
  \hline
  SRFBN~\cite{li2019feedback} & 0.373M & 3.217G & 46.05 & \underline{0.43} & \textbf{8.31} & \textbf{42.83} & 83.72 & 94.95 & 34.2945 & 0.9815 \\
  \hline
  HAN~\cite{niu2020single} & 64.342M & 495.0G & 45.70 & 0.39 & 7.46 & 40.90 & 83.61 & 96.17 & 36.6048 & 0.9826 \\
  \hline
  HiT-SNG~\cite{zhang2024hit} & 0.952M  & 13.324G & 45.01 & 0.39 & 7.34 & 40.19 & 81.98 & 95.17 & 35.4038 & 0.9805 \\
  \hline
  
  \multicolumn{9}{l}{\textit{Deep Unfolding}}  \\
  \hline
  ISTA-Net~\cite{CVPR2018ISTANet}  & 0.171M & 12.77G & 45.16 & 0.41 & 7.71 & 40.57 & 82.58 & 94.53 & 35.6733 & 0.9862 \\
  \hline
  ISTA-Net+~\cite{CVPR2018ISTANet}  & 0.337M & 24.33G & \underline{46.06} & 0.42 & 7.66 & 41.58& 84.46 & 96.17 & \textbf{38.4998} & \textbf{0.9887} \\
  \hline
  LAMP~\cite{Daubechies2004ISTA} & 2.126M & 0.278G & 14.22 & 0.05 & 1.11 & 7.31 & 21.56 & 41.06 & 27.8181 & 0.9405 \\
  \hline
  LIHT~\cite{Daubechies2004ISTA} & 21.10M & 1.358G & 10.35 & 0.06 & 0.92 & 4.99 & 14.74 & 30.50 & 27.5657 & 0.9198 \\
  \hline
  LISTA~\cite{Daubechies2004ISTA} & 21.10M & 1.358G & 30.13  & 0.25 & 4.13 & 22.29 & 51.18 & 72.82 & 30.1285 & 0.9537  \\
  \hline
  FISTA-Net~\cite{xiang2021fista} & 0.074M & 18.96G & 44.66 & \textbf{0.45} & 7.68 & 39.74 & 81.24 & 94.19& \underline{38.4773} & \underline{0.9872} \\
  \hline
  TiLISTA~\cite{Daubechies2004ISTA} & 2.126M  & 0.278G & 14.95 &0.06 & 1.23 & 7.72 & 22.50 & 46.23 & 27.7790 & 0.8895  \\
  \hline
  \rowcolor[rgb]{0.9,0.9,0.9} $\star$ \textbf{DISTA-Net (Ours)}  & 2.179M &35.10G   & \textbf{46.74}   & 0.38 &7.54 &\underline{42.44} &\textbf{86.18} & \textbf{97.14} & 38.3825 & \textbf{0.9887} \\
  \hline
  \end{tabular}
  \caption{Comparison with SOTA methods on the \textbf{CSIST-100K} dataset.}
  \label{tab:msar1}
  \vspace{-1\baselineskip}
  \end{table*}
  
\subsection{Experimental Settings} \label{subsec:setting}
%To validate the effectiveness of DISTA-Net, we conducted experiments using the CSIST-100K dataset.
% 为验证DISTA-Net由有效性，我们在数据集CSIST-100K进行了一些实验。
\textbf{Training.} Using CSIST-100K images as input, we apply Sec.~\ref{subsec:proximal}'s method to perform sub-pixel division with a sampling grid ratio of $c$. This generates an unmixed high-resolution grid as ground truth, where for each target $(x_i, y_i, g_i)$, the intensity $g_i$ is assigned to the pixel at position $\left(c \cdot x_i + \frac{c - 1}{2}, c \cdot y_i + \frac{c - 1}{2}\right)$ while other pixels remain zero. The selected $c$ value ensures each target appears as a distinct point, enabling accurate spatial separation.

\noindent \textbf{Testing.} We (1) apply post-processing with an intensity threshold of 50 to identify predicted targets; (2) project the unmixed grid back to the original $11 \times 11$ space via $\left(\frac{x_i-\left\lfloor\frac{c-1}{2}\right\rfloor}{c}, \frac{y_i-\left\lfloor\frac{c-1}{2}\right\rfloor}{c}\right)$. These mapped coordinates are then compared with ground truth target positions to compute CSO-mAP. Additionally, we calculate PSNR and SSIM by directly comparing the full predicted and ground truth high-resolution images.

\noindent \textbf{Hyperparameters.} 
% In our experiments, 
Our configuration uses: $c=3$ (baseline grid ratio), batch size of 64, DISTA-Net with 6 stages ($N=6$), and Dynamic Transform branch coefficient $(1-\alpha)=0.3$.

\subsection{Comparison with State-of-the-Art Methods} 
\noindent\textbf{Experimental Results.}
Table ~\ref{tab:msar1} compares DISTA-Net with traditional optimization (ISTA), image super-resolution, and deep unfolding methods on the CSIST-100K dataset, evaluated by computational efficiency (\#P/FLOPs), localization accuracy (CSO-mAP), and image quality (PSNR/SSIM).

For localization accuracy, we adopt CSO-mAP with different distance thresholds (from AP-05 to AP-25), where a smaller threshold indicates a stricter localization precision requirement. For instance, AP-05 evaluates the detection accuracy within 0.05-pixel distance, representing an extremely high precision demand. Considering the inherent error of 0.236 pixel width at subpixel division factor $c=3$, AP-20 and AP-25 (with 0.20 and 0.25 precision requirements respectively) serve as primary performance benchmarks. DISTA-Net achieves remarkable accuracy rates of 86.18\% and 97.14\% accuracy, outperforming most existing methods. On the mAP metric reflecting average CSIST unmixing performance, our method maintains a leading advantage with 46.74\% accuracy, demonstrating its robust localization capability across different precision requirements.

In terms of model efficiency, DISTA-Net achieves these results with moderate computational costs (2.179M parameters and 35.103G FLOPs), showing better efficiency compared to methods like ACTNet (46.212M, 62.798G) and HAN (64.342M, 0.495T). Additionally, DISTA-Net achieves a leading SSIM score of 0.9887 and a satisfactory PSNR of 38.3825, effectively preserving image details.

These comprehensive results validate that DISTA-Net achieves an effective balance between computational efficiency, localization accuracy, and reconstruction quality.

\begin{figure*}[htbp]
    \centering
    \includegraphics[width=.98\textwidth]{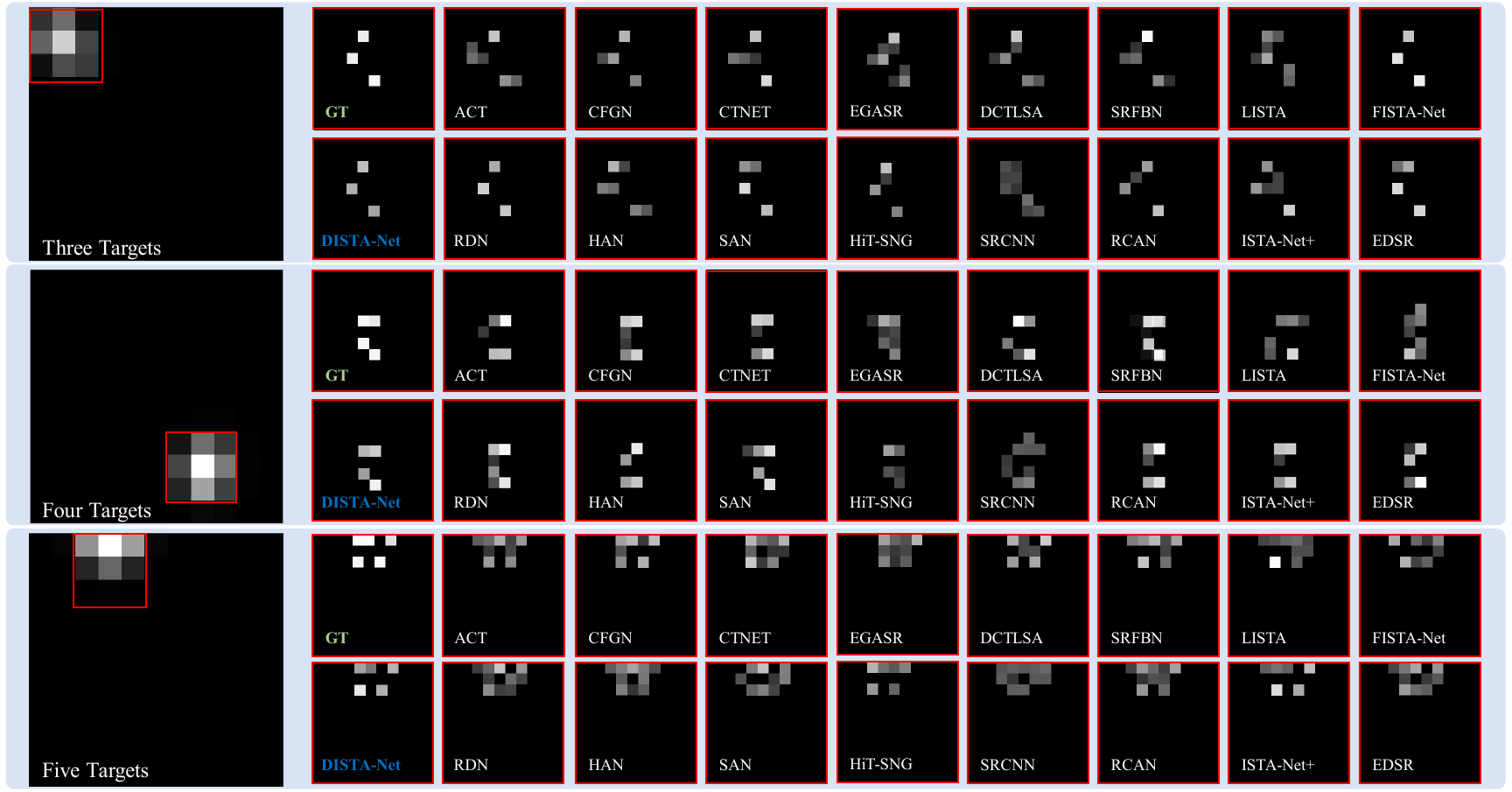}
    \caption{Visual comparison of $3 \times$ sub-pixel division reconstruction for scenes containing different numbers of closely-spaced infrared small targets. The red boxes highlight regions where targets exhibit significant sub-pixel characteristics.}
    \label{fig:compare}
\end{figure*}

\noindent\textbf{Visual comparison.} Fig.~\ref{fig:compare} compares reconstruction results with $3 \times$ sub-pixel division across methods for scenes containing $3\sim 5$ targets, assessing performance in dense multi-target scenarios. Existing methods struggle to reconstruct closely spaced targets, exhibiting blurred boundaries or merged detections, with degradation worsening for higher target counts (e.g., five-target cases). In contrast, DISTA-Net preserves both target quantity and sub-pixel positions while maintaining sharp boundaries and accurate spatial distributions, even under extreme density.

% 作为附录讨论不同超分倍率下的可视化以及阶段可视化分析
% \begin{figure*}[htbp]
%     \centering
%     \includegraphics[width=.99\textwidth]{./figs/experiment/stages_v2.pdf}
%     \caption{Stage-wise visualization comparison between FISTA-Net and DISTA-Net. (a) FISTA-Net demonstrates progressive sparsification across stages, directly reflecting the iterative shrinkage process. (b) DISTA-Net exhibits a two-phase reconstruction pattern, with initial feature enhancement (Stage 1-3) followed by target refinement (Stage 4-5), leading to more precise final reconstruction.}
%     \label{fig:v5}
% \end{figure*}
% Fig. ~\ref{fig:v5} compares the stage-by-stage visualization of FISTA-Net and DISTA-Net. FISTA-Net (Fig. X(a)) follows iterative shrinkage-thresholding, gradually increasing sparsity until bright pixels in the final stage show target locations. DISTA-Net (Fig. X(b)) uses a different approach: first enhancing target features through dynamic transform (Stages 1-3), then refining by suppressing false responses (Stages 4-5), achieving more precise target localization. This shows DISTA-Net learned a better reconstruction strategy, performing well in complex multi-target scenarios.

% These qualitative results further validate the effectiveness of our approach in handling complex multi-target scenarios, which is essential for practical applications requiring high-precision target detection and localization.

\label{subsec:sota}

\subsection{Ablation Study} \label{subsec:ablation}
\begin{table}[ht]
  \renewcommand\arraystretch{1.2}
  \footnotesize
  \centering
  %\vspace{-1\baselineskip}
  \vspace{-2pt}
  \resizebox{1\linewidth}{!}{
  \setlength{\tabcolsep}{2.pt}
  \begin{tabular}{l|c|ccccccc}
  {Method} & {CSO-mAP}  & AP-05  & AP-10    & AP-15    & AP-20  & AP-25 \\
  \Xhline{1pt}
  
  \hline
  ISTA-Net~\cite{CVPR2018ISTANet}    & 45.16 & \underline{0.41} & \underline{7.71} & 40.57 & 82.58 & 94.53 \\
  \hline
  DISTA-Net w/o DT  & 46.32 & 0.34 & 6.83 & 40.76& \underline{86.18} & \textbf{97.50}  \\
  \hline
  DISTA-Net w/o Thres.  & 46.17 & \textbf{0.44} & \textbf{7.77} & \underline{42.18} & 84.67 & 95.79  \\
  \hline
  \rowcolor[rgb]{0.9,0.9,0.9} $\star$ \textbf{DISTA-Net (Ours)}  & \textbf{46.74} &0.38 & 7.54 & \textbf{42.44} & \textbf{86.18} & \underline{97.14} \\
  \hline
  \end{tabular}
  }
  \caption{The effect of different components.}
  \label{tab:msar_2}
  \vspace{-0.5\baselineskip}
  \end{table}
\noindent \textbf{Effect of different components. }
We conduct ablation studies to evaluate the contribution of each component in DISTA-Net, with results shown in Table \ref{tab:msar_2}. The second row (DISTA-Net w/o DT) corresponds to the model variant without Dynamic Transform and the third row (DISTA-Net w/o Thres.) represents the model without Dynamic Soft-Thresholding. Compared to the baseline ISTA-Net, our complete model shows notable improvements in both CSO-mAP (45.16\% to 46.74\%) and AP-20 (82.58\% to 86.18\%).

Removing the Dynamic Transform leads to a slight performance decrease (CSO-mAP drops to 46.32\%), highlighting its role in enhancing the model's performance. The removal of Dynamic Soft-Thresholding results in the most significant performance degradation (CSO-mAP decreases to 46.17\%), emphasizing its key role in ensuring accuracy.

\begin{table}[htbp]
  \renewcommand\arraystretch{1.2}
  \footnotesize
  \centering
  %\vspace{-1\baselineskip}
  
  \vspace{-2pt}
  \resizebox{0.99\linewidth}{!}{
  \setlength{\tabcolsep}{6.pt}
  \begin{tabular}{l|c|c|c|c|c}
  
  \multirow{2}{*}{Method}   & \multirow{2}{*}{\#P $\downarrow$} & \multirow{2}{*}{FLOPs $\downarrow$} & \multicolumn{3}{c}{CSO-mAP}  \\
   & &  & mAP & AP-10    & AP-15  \\
   
  \Xhline{1pt}
  \multicolumn{4}{l}{\textit{c=5}}  \\
  \hline
  ISTA-Net ~\cite{CVPR2018ISTANet} &0.171M  & 39.544G   & 66.90 & 56.73 & 87.26 \\
  \hline
  ISTA-Net+ ~\cite{CVPR2018ISTANet}& 0.225M & 48.158G & \underline{68.50} & 57.96 & \underline{89.52}  \\
  \hline
  CFGN ~\cite{dai2023cfgn} & 0.538M & 4.122G & 67.95 & \underline{58.08} & 88.35 \\
  \hline
   \rowcolor[rgb]{0.9,0.9,0.9} $\star$ \textbf{DISTA-Net (Ours)} & 5.153M &102.4G & \textbf{69.58} &\textbf{60.95} & \textbf{90.85} \\
  \hline
  \multicolumn{4}{l}{\textit{c=7}}  \\
  \hline
  ISTA-Net ~\cite{CVPR2018ISTANet} & 0.171M & 89.51G & \underline{71.19} & \underline{76.45} & 84.16 \\
  \hline
  ISTA-Net+ ~\cite{CVPR2018ISTANet} & 0.225M & 103.0G & 71.09&74.90&\underline{84.90} \\
  \hline
  CFGN ~\cite{dai2023cfgn} & 0.548M & 4.202G & 70.38 & 73.88 & 83.97 \\
  \hline
  \rowcolor[rgb]{0.9,0.9,0.9} $\star$ \textbf{DISTA-Net (Ours)} & 6.409M & 142.3G & \textbf{72.84} &\textbf{78.47} & \textbf{86.09}  \\
  \hline
  \end{tabular}
  }
  \caption{Comparison of methods across different Sampling Grids on the \textbf{CSIST-100K} dataset.}
  \label{tab:msar_3}
  \vspace{-0.5\baselineskip}
  \end{table}
  
\noindent \textbf{Model Performance and Sampling Grid. }
% To investigate the impact of sampling grid ratio on model performance and computational efficiency, we conduct experiments with other different sampling grid ratios (c = 5, and 7) across various methods. Table~\ref{tab:msar_3} presents a comprehensive comparison of model performance under different sampling configurations.
We evaluate sampling ratios $c=5$ and $7$ across methods, analyzing their impact on performance and efficiency.
As the sampling grid ratio increases, all methods show improved detection performance due to better target positioning precision. DISTA-Net consistently outperforms other methods across all configurations. While ISTA-Net and ISTA-Net+ exhibit similar trends with lower overall performance, our proposed method achieves greater accuracy improvements for equivalent increases in sampling ratio. This advantage is particularly pronounced in the AP-10 ($c=5$) and AP-15 ($c=7$) metrics (see Table~\ref{tab:msar_3} for complete results). However, these performance gains are accompanied by substantially increased computational complexity. We recommend selecting an appropriate sampling grid ratio that aligns with specific application requirements to achieve an optimal balance between detection accuracy and computational efficiency.
% Notably, DISTA-Net demonstrates better efficiency in utilizing the increased resolution.

\noindent \textbf{Ours vs Super-Resolution + Detector Pipeline. }
The unmixing stage typically serves as a refinement step following the detection phase. For experimental rigor, we conducted experiments implementing the ``SR + Detector'' pipeline. We used YOLOv11 as the detector following leading SR methods (retrained on our IR data with unmixing GT for point-source super-resolution). The results demonstrate DISTA-Net's continued advantage: DISTA-Net + YOLOv11 achieved a CSO-mAP of 47.82, outperforming SRFBN + YOLOv11 (45.74) and CFGN + YOLOv11 (46.71). 
Visual analysis reveals that both conventional SR methods and our unmixing approach generate high-resolution images with well-resolved peaks, enabling effective target separation through simple thresholding. This demonstrates the dominant role of the unmixing stage in this task.

\noindent \textbf{Hyperparameters Analysis. }We analyzed model performance versus stage numbers and dynamic branch coefficients in \textbf{Supplementary}, demonstrating robust design.

\section{Conclusion} \label{sec:conclusion}

In this paper, we present the Dynamic Iterative Shrinkage Thresholding Network (DISTA-Net) to address CSIST unmixing task, which features adaptive generation of both convolution weights and thresholding parameters. Extensive experiments demonstrate that DISTA-Net achieves superior performance in both sub-pixel target detection accuracy and image reconstruction quality. To advance research in this domain, we introduce the CSIST dataset, CSO-mAP metric, and GrokCSO toolkit.

\noindent \textbf{Acknowledgement} This research is
supported by the NSFC (NO.62206133, 62301261, 62206134, U24A20330, 62361166670, 62225604) and the Shenzhen Science and Technology Program (JCYJ20240813114237048). Computation is supported by the Supercomputing Center of Nankai University.

% \clearpage

{
    \small
    \bibliographystyle{ieeenat_fullname}
    \bibliography{main}
}

\newpage

\end{document}